\documentclass[letterpaper, 10 pt, conference]{ieeeconf}  

\usepackage{amsmath,amsfonts}
\usepackage{algorithmic}
\usepackage{array}
\usepackage{textcomp}
\usepackage{pifont}
\usepackage{hyperref}
\usepackage{booktabs} 
\hypersetup{
	colorlinks=true, 
	linkcolor=blue,  
	citecolor=blue, 
	urlcolor=red     
}
\usepackage{url}
\usepackage{verbatim}
\usepackage{graphicx}
\usepackage{xcolor}

\def\BibTeX{{\rm B\kern-.05em{\sc i\kern-.025em b}\kern-.08em
		T\kern-.1667em\lower.7ex\hbox{E}\kern-.125emX}}
\usepackage[justification=centering]{caption}
\usepackage{times}
\usepackage{subcaption}
\usepackage{amsmath}
\usepackage{bm}
\usepackage{amsfonts}
\usepackage{amssymb}
\usepackage{booktabs}
\usepackage{float}
\usepackage{color}
\usepackage{ulem}
\usepackage{color, soul, framed}
\usepackage{bm}

\newtheorem{remark}{Remark}

\IEEEoverridecommandlockouts                              

\title{\LARGE \bf
Lightweight Kinematic and Static Modeling of Cable-Driven Continuum Robots via Actuation-Space Energy Formulation}

\author{Ke Wu$^{1}$, Yuhao Wang$^{1}$, Kevin Henry$^{1}$, Cesare Stefanini$^{1}$, Gang Zheng$^{2}$
}

\begin{document}

\maketitle
\thispagestyle{empty}
\pagestyle{empty}

\begin{abstract}
	Continuum robots, inspired by octopus arms and elephant trunks, combine dexterity with intrinsic compliance, making them well suited for unstructured and confined environments. Yet their continuously deformable morphology poses challenges for motion planning and control, calling for accurate but lightweight models. We propose the Lightweight Actuation-Space Energy Modeling (LASEM) framework for cable-driven continuum robots, which formulates actuation potential energy directly in actuation space. LASEM yields an analytical forward model derived from geometrically nonlinear beam and rod theories via Hamilton’s principle, while avoiding explicit modeling of cable–backbone contact. It accepts both force and displacement inputs, thereby unifying kinematic and static formulations. Assuming the friction is neglected, the framework generalizes to nonuniform geometries, arbitrary cable routings, distributed loading and axial extensibility, while remaining computationally efficient for real-time use. Numerical simulations validate its accuracy, and a semi-analytical iterative scheme is developed for inverse kinematics. To address discretization in practical robots, LASEM further reformulates the functional minimization as a numerical optimization, which also naturally incorporates cable potential energy without explicit contact modeling.

\end{abstract}


\section{INTRODUCTION}

Continuum robots, inspired by biological structures such as octopus arms and elephant trunks, possess continuously deformable bodies with theoretically infinite degrees of freedom (DoFs) \cite{walker2013continuous}. This morphology endows them with exceptional dexterity and compliance in unstructured and confined environments \cite{russo2023continuum}, enabling applications in minimally invasive surgery \cite{burgner2015continuum}, safe manipulation \cite{russo2023continuum}, human–robot interaction \cite{abah2021multi}, and exploration \cite{wooten2018exploration}. Realizing these capabilities, however, requires accurate and computationally efficient modeling \cite{webster2010design}, which underpins motion planning \cite{russo2023continuum} and control \cite{george2018control}.

Over the past two decades, diverse modeling approaches have been developed for continuum robots, including piecewise constant curvature (PCC) models \cite{webster2010design}\cite{chirikjian1992theory}, pseudo-rigid-body approximations \cite{wang2025spirobs}, geometrically nonlinear beam and rod theories \cite{till2017elastic}\cite{tummers2023cosserat}, and FEM-based methods \cite{coevoet2017software}. Each entails different trade-offs in fidelity, generality, and computational efficiency, and the choice largely depends on application demands and real-time feasibility \cite{burgner2015continuum}\cite{webster2010design}\cite{jones2006kinematics}.


The piecewise constant curvature (PCC) model and its variants are widely used assumption-based kinematic approaches, appreciated for their simplicity and compatibility with rigid-link formulations such as Denavit–Hartenberg \cite{jones2006kinematics}\cite{chirikjian1992theory}. They offer closed-form solutions with low computational cost, making them suitable for real-time control and planning \cite{burgner2015continuum}. However, their reliance on geometric assumptions limits generality, and accuracy deteriorates under external loading or gravity \cite{russo2023continuum}.


In contrast, physics-based models such as Cosserat rods \cite{tummers2023cosserat} and FEM \cite{coevoet2017software} offer greater generality and physical accuracy, capturing nonlinear deformations, complex actuation, and constraints. However, they often fail to meet real-time requirements without model reduction or simplification \cite{webster2010design}\cite{goury2018fast}\cite{mathew2025reduced}, which compromises fidelity. In the meanwhile, from a strain-mode perspective, classical PCC models and Euler–Bernoulli beam theory primarily capture bending \cite{jones2006kinematics}. PCC has been extended to include axial elongation \cite{jones2006kinematics}, while Kirchhoff beams incorporate bi-axial bending and torsion but neglect axial stretch \cite{till2017elastic}. Cosserat rod models capture all four essential strain modes including biaxial bending, torsion, and axial extensibility, thus fully representing spatial continuum deformation \cite{tummers2023cosserat}. FEM further supports these modes and accommodates complex geometries and loads \cite{coevoet2017software}. Broader strain representation enhances fidelity, particularly for robots with intricate shapes, nonuniform loads, or diverse actuation \cite{armanini2023soft}.


Another major challenge is formulating inverse models, which underpins motion planning, trajectory generation, and feedforward control \cite{della2023model}. Under the PCC assumption, Neppalli \cite{neppalli2008geometrical} derived closed-form inverse kinematics, and more recently Dickson \cite{dickson2025real} demonstrated that PCC-based dynamics are differentially flat, enabling analytical inverse solutions and trajectory generation up to 23× faster than real time. To date, PCC remains the only modeling class proven differential flatness. In contrast, physics-based models still rely on iterative solvers, limiting real-time use \cite{della2023model}. 

Beyond accuracy and generality, model complexity and ease of implementation strongly affect practical adoption \cite{webster2010design}. PCC \cite{jones2006kinematics} and pseudo-rigid-body models \cite{wang2025spirobs} are favored for their simplicity and compatibility with rigid-body frameworks, easing integration into simulators such as MuJoCo \cite{wang2025spirobs}. Physics-based methods, such as Cosserat rods \cite{tummers2023cosserat} and FEM \cite{coevoet2017software}, provide higher fidelity and accommodate complex deformations and actuation, but their substantial mathematical and computational complexity limits wider adoption compared to the case of PCC.


\begin{figure}[]
	\vspace*{0.3cm} 
	\begin{centering}
		\includegraphics[trim=0 0 0 0, clip, scale=1.1]{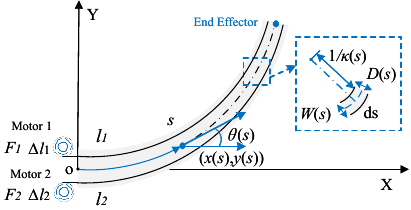}
		\caption{Schematic Diagram of a Planar Continuum Robot}
		\label{Schematic diagram}
	\end{centering}
\end{figure}

Given the trade-offs between simplified analytical models and complex physics-based approaches, this work introduces the Lightweight Actuation-Space Energy Modeling (LASEM) framework for cable-driven continuum robots. LASEM formulates actuation potential energy directly in actuation space, yielding an analytical forward model derived from geometrically nonlinear beam and rod theories via Hamilton’s principle, while avoiding explicit modeling of cable–backbone contact reported in many previous methods in the literature \cite{tummers2023cosserat}\cite{renda2014dynamic}. It accommodates both force and displacement inputs, thereby unifying kinematic and static formulations. Assuming the friction is neglected, the framework generalizes to arbitrary cable routings, nonuniform geometries, distributed loading and axial extensibility, while remaining lightweight and computationally efficient for real-time applications. Numerical simulations by classic approaches validate the model, and a semi-analytical iterative scheme is introduced for inverse kinematics. To bridge theory and practice, discretization effects are addressed by reformulating the functional minimization as a numerical optimization, which also naturally incorporates cable potential energy without explicit contact modeling.

\section{Analytical Forward-Inverse Kinematics \\ via LASEM Framework}\label{FIK}

This section presents LASEM framework through a systematic derivation. The studied continuum robot (Fig.~\ref{Schematic diagram}) has varying cross-section $D(s)$ and cable spacing $W(s)$, with two inextensible cables symmetrically embedded and independently driven by base-mounted motors. Cable lengths along the robot body are $l_1, l_2$, with corresponding forces $F_1, F_2$ and displacements $\Delta l_1, \Delta l_2$. The backbone is parameterized by arc length $s \in [0,L]$, with position $(x(s),y(s))$ and tangent angle $\theta(s)$. The curvature is $\kappa(s) = \frac{\mathrm{d}\theta}{\mathrm{d}s}$, where $1/\kappa(s)$ gives the local radius of curvature. For clarity, we first consider a representative case with constant $D$ and $W$, i.e., a uniform cross-section with parallel, symmetrically placed cables.


\subsection{Geometric Relationships and Energy Formulations}
In this section, we derive the robot’s energy formulations within a Lagrangian framework, incorporating the geometric constraints of cable-driven actuation. As shown in Fig.~\ref{Schematic diagram}, the two inextensible cables constrain deformation by enforcing geometric relations with the backbone. Their local curvatures, offset laterally from the backbone and thus differing from it, are expressed as:
\begin{equation}
	\small
		\begin{aligned}
	\kappa_1(s) = \frac{1}{\frac{1}{\kappa(s)} - \frac{W}{2}}, \quad
	\kappa_2(s) = \frac{1}{\frac{1}{\kappa(s)} + \frac{W}{2}}
			\end{aligned}
\end{equation}
where $\kappa$ is the backbone curvature of the continuum robot at $s$. Then, we have the following relationship for the cable 
lengths $l_1$ and $l_2$: 
\begin{equation}
	\small
	\begin{aligned}
		l_{1,2} & = \int_{0}^{\theta(L)}\!\!\!\!\!\!\! \frac{1}{\kappa_{1,2}(s)} \, \mathrm{d}\theta 
		= L \pm \frac{W}{2} \left( \theta(L) - \theta(0) \right) 
		= L \pm \frac{\theta(L)}{2} W \\
	\end{aligned}
\end{equation}
Clearly, we can also define the displacement of each cable:
\begin{equation}\label{deltal}
	\small
\begin{aligned}	
\Delta l=l_1-L=\Delta l_1=L-l_2=-\Delta l_2 = \frac{\theta(L)}{2} W
		\end{aligned}
\end{equation}
Next, we can formulate the total potential energy of the studied continuum robot using geometrically nonlinear Euler–Bernoulli beam theory:
\begin{equation}\label{energy}
	\small
		\begin{aligned}
	\Pi = E_p - W_p &= \int_0^L \frac{1}{2} EI \left( \frac{\mathrm{d}\theta}{\mathrm{d}s} \right)^2 \mathrm{d}s - F_1 \Delta l_1  - F_2 \Delta l_2\\
	&=\int_0^L \frac{1}{2} EI \left( \frac{\mathrm{d}\theta}{\mathrm{d}s} \right)^2 \mathrm{d}s - \Delta_F \Delta l
		\end{aligned}
\end{equation}
where only the bending strain $\tfrac{d\theta}{ds}$ is considered. The cable force difference is defined as $\Delta_F = F_1 - F_2$, while $E$ and $I$ denote the Young’s modulus and the cross-sectional second moment of area. Crucially, the terms $- F_1 \Delta l_1  - F_2 \Delta l_2$ or $-\Delta_F \Delta l$ represent the actuation potential energy expressed directly in actuation space, which forms the basis for deriving the subsequent lightweight analytical solution. Based on the principle of minimum potential energy, together with equations \eqref{deltal} and \eqref{energy}, the statics-based kinematics can be formulated as the following functional minimization problem:	
\begin{equation}\label{Pi}
	\small
	\begin{aligned}
	\min_{\theta(s)} \; 	\Pi = \int_0^L \frac{1}{2} EI \left( \frac{\mathrm{d}\theta}{\mathrm{d}s} \right)^2 \mathrm{d}s - \frac{1}{2} \Delta_F W \theta(L)
	\end{aligned}
\end{equation}
The function $\theta(s)$ that minimizes the total potential energy $\Pi$ corresponds to the static equilibrium configuration of the continuum robot under the applied differential force input $\Delta_F$, subject to the boundary condition $\theta(0) = 0$.
\subsection{Analytical Solutions through Functional Variation}\label{derivation1}
\begin{figure}[]
	\vspace*{0.3cm} 
\centering	
\begin{subfigure}[b]{1.0\linewidth}
	\centering
	\includegraphics[trim=0 0 0 0, clip, width=1.\linewidth]{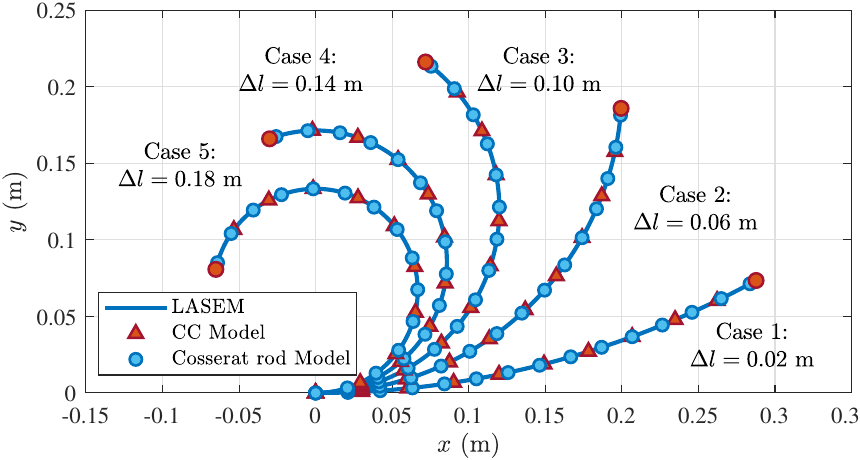}
	\caption{Forward Model Verification}
	\label{FMV}
\end{subfigure}
\begin{subfigure}[b]{1.0\linewidth}
	\centering
	\includegraphics[trim=0 30 20 30, clip, width=1\linewidth]{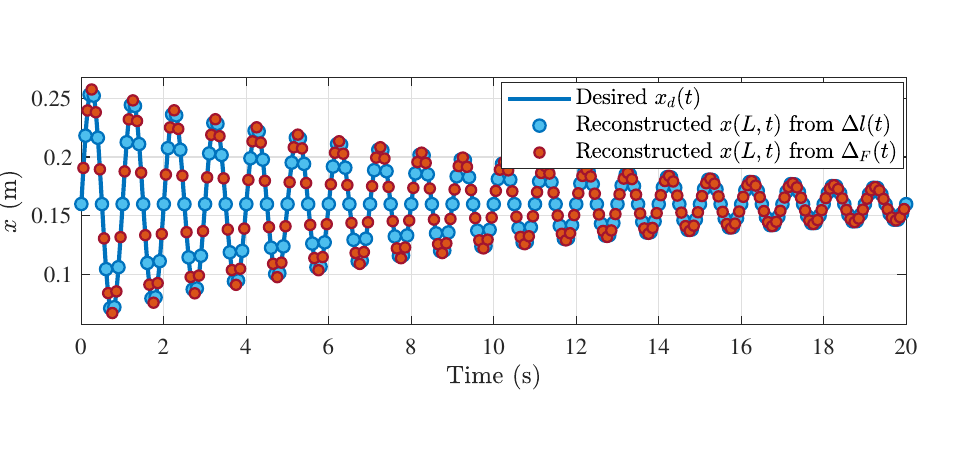}
	\caption{Inverse Model Verification}
	\label{IKV}
\end{subfigure}
\caption{Numerical Verification - Classic Case}
\label{CCr}
\end{figure}

To find the function $\theta(s)$ that minimizes the total potential energy $\Pi$, we require the first variation $\delta \Pi = 0$ to hold for all admissible perturbations $\delta\theta(s)$. First, we apply a small perturbation $\delta \theta(s)$:
\begin{equation}
\small
	\begin{aligned}
	\delta \Pi = \Pi[\theta(s) + \delta \theta(s)] - \Pi[\theta(s)]
\end{aligned}
\end{equation}
Keeping only first-order terms not only due to the infinitesimal magnitude of $\delta \theta(s)$, but also because of the small variations of perturbations along the structure, we can neglect the squared term:
\begin{equation}\label{pi}
			\small
	\begin{aligned}
	\delta \Pi = \int_0^L EI \frac{\mathrm{d}\theta}{\mathrm{d}s} \frac{\mathrm{d}(\delta \theta)}{\mathrm{d}s} \mathrm{d}s - \frac{1}{2} \Delta_F W \delta \theta(L)
\end{aligned}
\end{equation}
The first term is integrated by parts to facilitate variational analysis:
\begin{equation}\label{parts}
			\small
	\begin{aligned}
	\int_0^L EI  \frac{\mathrm{d}\theta}{\mathrm{d}s} \frac{\mathrm{d}(\delta \theta)}{\mathrm{d}s} \mathrm{d}s = \left. EI \frac{\mathrm{d}\theta}{\mathrm{d}s}\delta \theta(s) \right|_0^L - \int_0^L EI \frac{\mathrm{d}^2 \theta}{\mathrm{d}s^2}	 \delta \theta(s) \mathrm{d}s
\end{aligned}
\end{equation}
Then, we can rearrange \eqref{pi} through \eqref{parts}:
\begin{equation}\label{key}
\small
\begin{aligned}	
	\delta \Pi = \left. EI \frac{\mathrm{d}\theta}{\mathrm{d}s}\delta \theta(s) \right|_0^L - \int_0^L EI \frac{\mathrm{d}^2 \theta}{\mathrm{d}s^2}	 \delta \theta(s) \mathrm{d}s - \frac{1}{2} \Delta_F W \delta \theta(L)	
\end{aligned}
\end{equation}
From Eq.\eqref{key}, we observe that $\delta \Pi$ contains two types of terms: one multiplied by $\delta\theta(s)$ and another by $\delta\theta(L)$. To ensure $\delta \Pi = 0$ for arbitrary variations $\delta\theta(s)$, the coefficient of $\delta\theta(s)$ in the integrand must vanish, yielding the differential equation:
\begin{equation}\label{1}
\small
	\begin{aligned}
	EI \frac{\mathrm{d}^2\theta}{\mathrm{d}s^2} = 0
\end{aligned}
\end{equation}
Meanwhile, the boundary term involving $\delta\theta(L)$ must also vanish. Collecting the terms involving $\delta\theta(L)$ in Eq. \ref{key}, we obtain:
\begin{equation}\label{2}
\small
	\begin{aligned}
	EI \frac{\mathrm{d}\theta}{\mathrm{d}s} \delta \theta(s) \big|_0^L - \frac{1}{2} \Delta_F W \delta \theta(L) = 0
	\footnotesize
	\end{aligned}
\end{equation}
Obviously, we can derive this analytical static model for the studied continuum robot in Fig. \ref{Schematic diagram} given \eqref{1} and \eqref{2}:
\begin{equation}\label{static}
	\small
	\begin{aligned}
	\theta(s) = \frac{W \Delta_F}{2EI} s
		\end{aligned}
\end{equation}
Meanwhile, it is also straightforward to derive the following statics-based kinematic model using the boundary condition at $s=L$ of \eqref{static} and the geometric constraint \eqref{deltal}:
\begin{equation}\label{kine}
	\small
				\begin{aligned}
	\theta(s) = \frac{2 \Delta l}{WL} s
			\end{aligned}
\end{equation}
The result shows that $\theta(s)$ varies linearly with arc length $s$, yielding a constant curvature $\kappa(s) = \tfrac{d\theta}{ds}$. This physically confirms the constant-curvature assumption of the PCC model \cite{chirikjian1992theory}\cite{jones2006kinematics}. The formulation accommodates both force and displacement control by taking either the cable force difference $\Delta_F$ or displacement difference $\Delta l$ as input. Moreover, from \eqref{static} and \eqref{kine}, their explicit relationship follows as:
\begin{equation}\label{lf}
	\small	
	\begin{aligned}
	\Delta_F =\frac{4EI}{W^2L}\Delta l
	\end{aligned}
\end{equation}
This indicates a linear relation between $\Delta l(t)$ and $\Delta_F$ under negligible friction, which can be exploited for motion planning and control in continuum robots. Building on the solution of $\theta(s)$, the backbone coordinates admit closed-form expressions:
\begin{equation}\label{xy}
	\small	
	\begin{aligned}
		x(s)=\int_0^s \cos\theta(\xi) \mathrm{d}\xi;\ y(s)=\int_0^s \sin\theta(\xi) \mathrm{d}\xi
	\end{aligned}
\end{equation}
The proposed LASEM expressions \eqref{static} to \eqref{xy} provide a lightweight and efficient way to compute robot poses, improving interpretability and avoiding iterative solvers. As shown in Fig.~\ref{FMV}, LASEM not only recovers the CC model \cite{jones2006kinematics} but also matches the Cosserat rod model solved numerically \cite{mathew2022sorosim} using the force input from \eqref{lf}.
\begin{table}[t]
	\vspace*{0.3cm} 
	\centering
	\caption{Geometric and material parameters}	\small
	\label{tab:parameters}
	\begin{tabular}{llcc}
		\toprule
		\textbf{Parameter}         & \textbf{Symbol}       & \textbf{Value} & \textbf{Unit} \\
		\midrule
		Rod length                 & $L$                   & $0.3$          & m             \\
		Rod diameter               & $D$                   & $0.004$& m             \\
		Young's modulus            & $E$                   & $2 \times 10^9$& Pa           \\
		Cable spacing             & $W$                   & $0.11$& m             \\
		Second moment of area      & $I = \frac{\pi}{64}D^4$ & $1.26 \times 10^{-11}$& m$^4$ \\
		Cross-section area      & $A = \frac{\pi}{4}D^2$ & $1.26 \times 10^{-5}$& m$^2$ \\
		\bottomrule
	\end{tabular}
\end{table}
\subsection{Semi-analytical Inverse Kinematic Formulations}
Inverse kinematics is crucial for motion planning and control of continuum robots but remains challenging due to their nonlinear and infinite-dimensional nature \cite{russo2023continuum}. Closed-form solutions are especially valuable for real-time use. In this work, actuation is represented by a single input, either cable displacement $\Delta l$ or force difference $\Delta_F$, while the tip coordinate $x(L)$ is chosen as the system output. A time-varying constraint is then imposed on $x(L)$ using the closed-form forward model from \eqref{static}:
\begin{equation}
	\small	
	\begin{aligned}
		x(L,t) = \int_0^L \cos\left(\frac{ W\Delta_F(t)}{2EI}s \right) \,\mathrm{d}s
	\end{aligned}
\end{equation}
Differentiating both sides with respect to time yields:
\begin{equation}\label{xdot}
	\small			
	\begin{aligned}
		\dot{x}(L,t) = - \dot{\Delta}_F(t)\int_0^L \sin\left( \frac{W\Delta_F(t)}{2EI}s \right) \left( \frac{W}{2EI}s \right) \, \mathrm{d}s
	\end{aligned}
\end{equation}
Rearranging gives a semi-analytical formulation for solving the inverse statics input:
\begin{equation}
	\small		
	\begin{aligned}
		\dot{\Delta}_F(t) = \frac{-\dot{x}(L,t)}{\int_0^L \sin\left( \frac{W\Delta_F(t)}{2EI}s \right)\left( \frac{W}{2EI}s \right) \, \mathrm{d}s}
	\end{aligned}
\end{equation}
Given a desired trajectory $x(L,t)$, this formulation allows us to explicitly iteratively compute the required actuation force $\Delta_F(t)$ using numerical schemes such as explicit Euler integration. Similarly, a parallel formulation can be derived for inverse statics-based kinematics, where the actuation input is instead expressed in terms of the displacement input $\Delta l$:
\begin{equation}\label{xd}
	\small			
	\begin{aligned}
		\dot{\Delta} l(t) = \frac{-\dot{x}(L,t)}{\int_0^L \sin\left( \frac{2\Delta l(t)}{WL}s\right)\left(\frac{2s}{WL}\right) \mathrm{d}s}
	\end{aligned}
\end{equation}
\begin{remark}
	Both \eqref{static} and \eqref{kine} depend on $\dot{x}(L,t)$ rather than $x(L,t)$, so the initial input $\Delta_F(0)$ or $\Delta l(0)$ must be consistent with $x(L,0)$. Moreover, the denominators may vanish in certain cases, introducing singularities that should be avoided or mitigated using methods such as Damped Least Squares \cite{buss2005selectively}.
\end{remark}
As graphically illustrated in Fig.~\ref{Schematic diagram}, the mechanical and geometric properties of the studied continuum robot are summarized in Table~\ref{tab:parameters}. Under this setting, Fig.~\ref{IKV} demonstrates that the inverse model generates the input that enables the end-effector position $x(L,t)$ to accurately follow the desired trajectory:
\begin{equation}\label{xd1}
	\small			
	\begin{aligned}
	x_d(t) = x_0 + 0.1 e^{-0.1 t} \sin(2\pi t),
	\end{aligned}
\end{equation}
where $x_0$ corresponding to the initial displacement input $\Delta l(0)=0.1$ m.
\section{Generalized Analytical Kinematics with LASEM: Case Studies}
Building on the constant-curvature case, we generalize LASEM framework to capture nonlinear deformations induced by structural and actuation complexities, such as arbitrary cable routings, nonuniform geometries, distributed loading, axial extensibility. These effects are consistently integrated within a unified functional-minimization formulation, forming the generalized modeling capability of the LASEM framework.
\begin{equation}\label{general}
			\small
			\begin{aligned}
	\min_{\theta(s)} \; 	\Pi =& \int_0^L \!\! [\frac{1}{2} EI(s) ( \frac{\mathrm{d}\theta}{\mathrm{d}s} )^2 \!\! \!+ \frac{1}{2} EA(s) u(s)^2] \mathrm{d}s - F_1 \Delta l_1\\ &- F_2 \Delta l_2
	- \int_0^L q_x (s - x(s)) \mathrm{d}s + \int_0^L q_y y(s) \mathrm{d}s
\end{aligned}
\end{equation}
Here, $A(s)$ and $u(s)$ denote the cross-sectional area and axial strain, while $q_x$ and $q_y$ represent distributed loads such as gravity. Since the geometrically nonlinear Euler--Bernoulli beam theory neglects axial strain, a reduced-mode Cosserat rod model is employed to capture this effect \cite{wu2023design}. Representative case studies are presented based on this generalized model. As the analytical derivations largely parallel those in Section~\ref{FIK}, only the final closed-form solutions are reported, with deviations from the standard procedure explicitly noted where applicable.
\begin{figure}[]
	\vspace*{0.3cm} 
	\centering	
	\begin{subfigure}[b]{1.0\linewidth}
		\centering
		\includegraphics[trim=0 0 0 0, clip, width=0.8\linewidth]{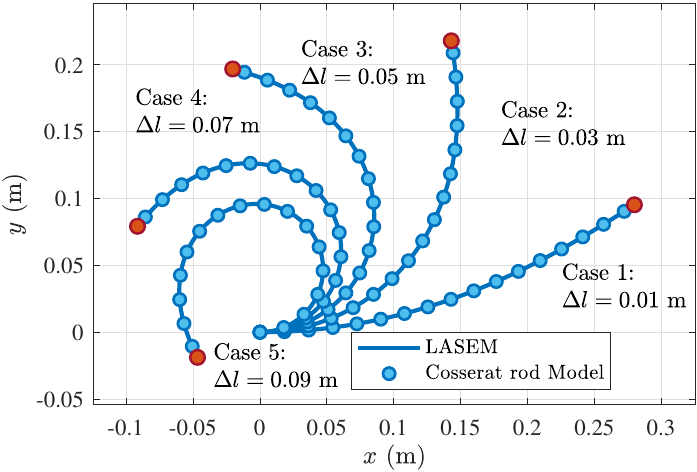}
		\caption{Forward Model Verification}
		\label{FMV1}
	\end{subfigure}
	\begin{subfigure}[b]{1.0\linewidth}
		\centering
		\includegraphics[trim=15 15 0 30, clip, width=1\linewidth]{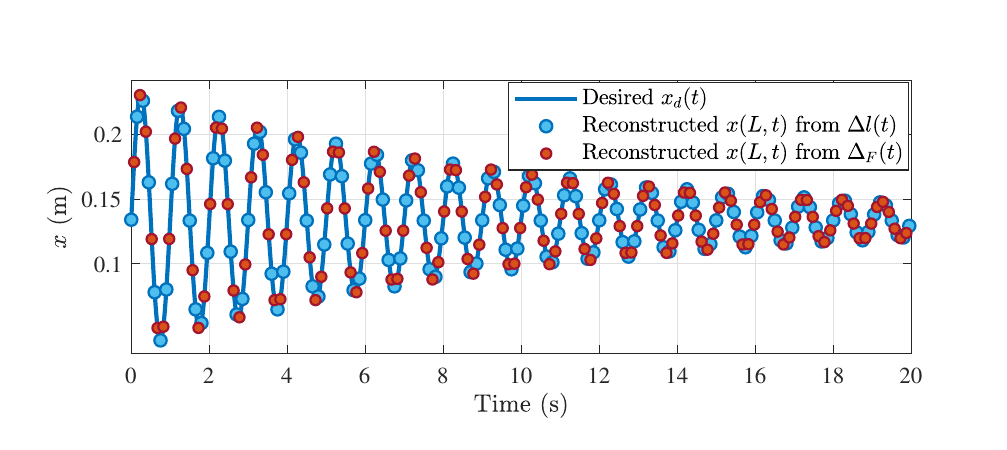}
		\caption{Inverse Model Verification}
		\label{IKV1}
	\end{subfigure}
	\caption{Numerical Verification - Arbitrary cable routings}
	\label{Wsr}
\end{figure}
\subsubsection{Arbitrary Cable Routings} \label{VCP}
Cable routing critically influences the kinematic response of continuum robots, shaping both workspace and actuation behavior \cite{renda2022geometrically}. Closed-form solutions for arbitrary cable routing $W(s)$ provide useful guidance for design and control. Here, two cables are symmetrically placed about the centerline with lateral spacing, as shown in Fig.~\ref{Schematic diagram}. Using a variational formulation analogous to Section~\ref{FIK}, we derive the following analytical expression for $\theta(s)$, which explicitly captures the effect of arbitrary cable routing on the robot’s deformed shape:
\begin{equation}
			\small	
	\begin{aligned}
	\theta(s) = \frac{\Delta_F}{2EI} \int_0^s W(\xi) \, \mathrm{d}\xi = \frac{2\Delta l}{\int_0^L W(\sigma)^2 \, \mathrm{d}\sigma} \int_0^s W(\xi) \, \mathrm{d}\xi
\end{aligned}
\end{equation}
with the relationship between $\Delta_F$ and $\Delta l$:
\begin{equation}\label{111}
		\small	
	\begin{aligned}
	\Delta_F = \frac{4EI}{\int_0^L W(s)^2 \, \mathrm{d}s}\Delta l
	\end{aligned}
\end{equation}
and the semi-analytical inverse kinematic formulations are as follows:
\begin{equation}
	\small		
	\begin{aligned}
		&\dot{\Delta} l(t) = \frac{-\dot{x}(L,t)}{\frac{2}{\int_0^L W(\sigma)^2 \mathrm{d}\sigma}\int_0^L \sin(\theta(s,t))\left( \int_0^s W(\xi) \, \mathrm{d}\xi \right) \mathrm{d}s}\\
		&\dot{\Delta}_F(t) = \frac{-\dot{x}(L,t)}{\frac{1}{2EI} \int_0^L \sin(\theta(s,t))\left( \int_0^s W(\xi) \, \mathrm{d}\xi \right) \mathrm{d}s}\\
	\end{aligned}
\end{equation}
To study the effect of nonuniform cable routing, we define a spatially varying cable spacing as:
\begin{equation}	\small	
	W(s) = 0.04 - 0.03 \left( \frac{s}{L} \right)^3.
\end{equation}
The remaining parameters are kept consistent with Table~\ref{tab:parameters}. For the inverse model verification, the trajectory given by \eqref{xd1} and $\Delta l(0)=0.1\ $m is used as the tracking reference. The corresponding numerical verification of the forward and inverse models are shown in Fig. \ref{Wsr}.
\begin{figure}[]
	\vspace*{0.3cm} 
	\centering	
	\begin{subfigure}[b]{1.0\linewidth}
		\centering
		\includegraphics[trim=0 0 0 0, clip, width=0.98\linewidth]{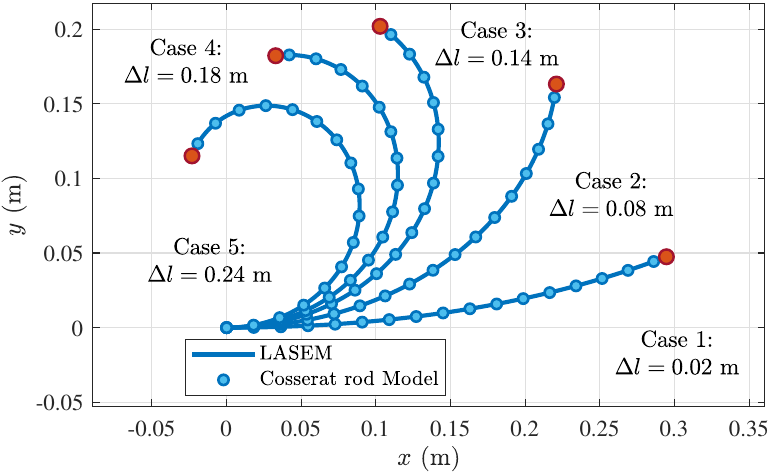}
		\caption{Forward Model Verification}
		\label{FMV2}
	\end{subfigure}
	\begin{subfigure}[b]{1.0\linewidth}
		\centering
		\includegraphics[trim=15 30 0 40, clip, width=1.1\linewidth]{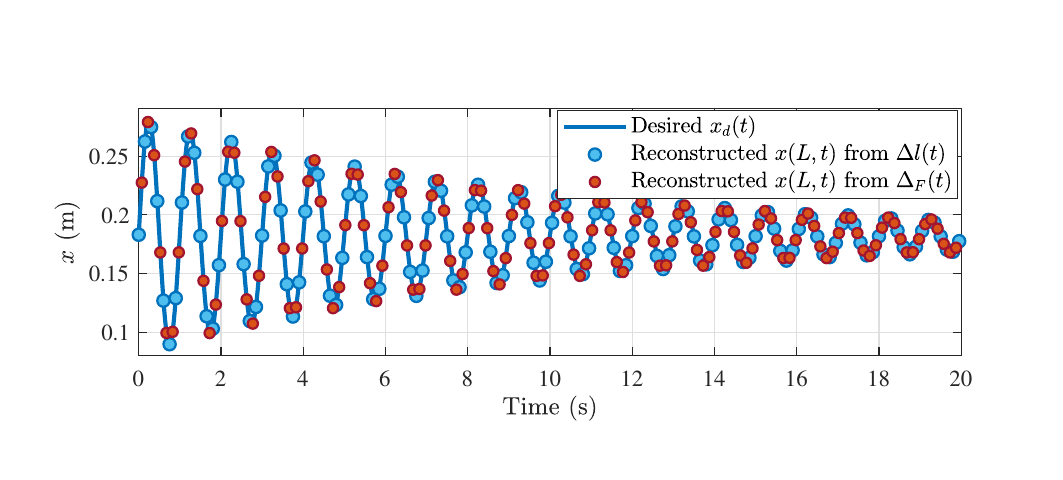}
		\caption{Inverse Model Verification}
		\label{IKV2}
	\end{subfigure}
	\caption{Numerical Verification - Arbitrary Cable Routings}
	\label{VCSr}
\end{figure}
\subsubsection{Nonuniform Geometries}\label{VCS}
Variations in cross-sectional geometry alter the bending stiffness and consequently the deformation of continuum robots. The following analytical expressions describe curvature as a function of the spatially varying flexural rigidity $EI(s)$:
\begin{equation}\label{VCSe}
			\small
	\begin{aligned}
	\theta(s) = \frac{W\Delta_F }{2}\beta(s)=\frac{2\Delta l}{W}\frac{\beta(s)}{\beta(L)}
\end{aligned}
\end{equation}
where $\beta(s)$ is a parameter characterizing the overall geometry and stiffness distribution of the continuum robot:
\begin{equation}
		\small
	\begin{aligned}
	\beta(s)= \int_0^s \frac{1}{EI(\xi)} \, \mathrm{d}\xi
\end{aligned}	
\end{equation}
Likewise, the relationship between $\Delta_F$ and $\Delta l$ can be derived from \eqref{VCSe}:	
 \begin{equation}\label{222}
 	\Delta_F = \frac{4}{W^2\beta(L)}\Delta l
 \end{equation}
The closed-form inverse kinematic solutions are presented below:
\begin{equation}
	\small	
	\begin{aligned}
		&\dot{\Delta}_F(t) = \frac{-\dot{x}(L,t)}{ \frac{W}{2} \int_0^L \sin\left( \theta(s,t)\right)\beta(s) \mathrm{d}s }\\
		&\dot{\Delta} l(t) = 
		\frac{-\dot{x}(L,t)}
		{\frac{2}{W \beta(L)} \int_0^L \sin\left( \theta(s,t)\right)\beta(s) \mathrm{d}s}
	\end{aligned}
\end{equation}
Here, we choose a nonlinear varying second moment of area to reflect a tapering rod structure. Specifically, the diameter decreases linearly from base to tip, and the corresponding second moment of area is given by 
\begin{equation}\small
	I(s) = \frac{\pi}{64} \left(D_0 + (D_1 - D_0)\tfrac{s}{L}\right)^4
\end{equation}
where $D_0 = 0.006\ \mathrm{m}$ and $D_1 = 0.005\ \mathrm{m}$ are the base and tip diameters, respectively. Other parameters follow Table~\ref{tab:parameters}. For inverse-model verification, the trajectory in \eqref{xd1} with $\Delta l(0)=0.1\ \mathrm{m}$ is used as reference. Numerical results of the forward and inverse models are shown in Fig.~\ref{VCSr}.
\subsubsection{Distributed Loading}\label{DL}
Distributed loading, particularly gravity, significantly affects continuum robot performance. Although a analytical closed-form solution is not attainable here following Section~\ref{FIK}, the governing differential equations and boundary conditions explicitly incorporate the distributed forces $q_x$ and $q_y$ along the robot body:
\begin{equation}\label{bvp}
	\small
	\begin{aligned}
		&\textbf{D.E.}\ 	EI\frac{\mathrm{d}^2 \theta}{\mathrm{d}s^2} = - q_x (L - s) \sin(\theta(s)) + q_y (L - s) \cos(\theta(s))\\
		&\textbf{B.C.}\ 	\theta(0) = 0,\ \frac{\mathrm{d} \theta}{\mathrm{d}s}(L) = \frac{W \Delta_F}{2EI}
	\end{aligned}	
\end{equation}
\begin{figure}[]
	\vspace*{0.3cm} 
	\begin{centering}
		\includegraphics[trim=0 0 0 0, clip, width=1.02\linewidth]{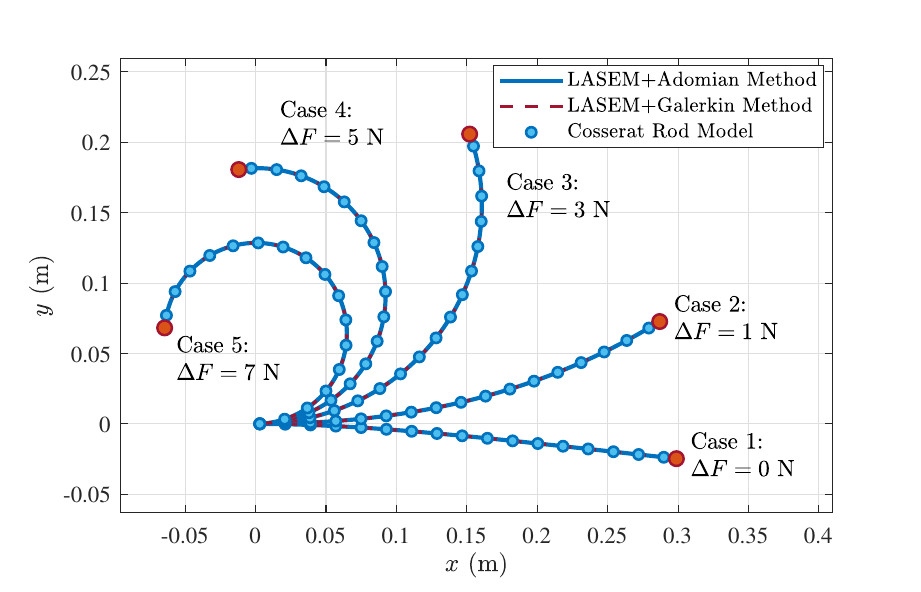}
		\caption{Numerical Verification - Distributed Loading}
		\label{DLr}
	\end{centering}	
\end{figure}Importantly, different to the classic modeling methods \cite{tummers2023cosserat}\cite{renda2014dynamic}, the formulation avoids the need to explicitly model contact forces between the cables and the structure, greatly simplifying the modeling process. 
The above resulting equations are amenable to efficient numerical methods, such as Galerkin projection:	
\begin{equation}
	\small
	\begin{aligned}
	\int_0^L \mathcal{W}(s)\mathcal{R}(s) \mathrm{d}s=0
	\end{aligned}	
\end{equation}
where $\mathcal{W}(s)$ denotes the Galerkin weight function and $\mathcal{R}(s)$ denotes the residual function derived from the D.E. in \eqref{bvp}. Otherwise, we can also choose approximate analytical techniques like the Adomian decomposition method \cite{wu2023design}:	
\begin{equation}\label{AMD}
	\small
	\begin{aligned}
		&\mathcal{L}[{\theta}({s})]+\mathcal{N}[{\theta}({s})]=0
	\end{aligned}	
\end{equation}
where $\mathcal{L}[.]$ here denotes the second derivative operator $\frac{\mathrm{d}^2}{\mathrm{d}s^2}[.]$ and $\mathcal{N}[.]$ denotes the nonlinear operator:	
\begin{equation}\label{AMD1}
	\small
	\begin{aligned}
		&\mathcal{N}[\theta(s)]= \frac{q_x (L - s) \sin(\theta(s)) - q_y (L - s) \cos(\theta(s))}{EI}
	\end{aligned}	
\end{equation}
Integrating \eqref{AMD} twice (equivalently, $\mathcal{L}[.]^{-1}$) with B.C. satified, we can arrive at its semi-analytical closed-form solution:
\begin{equation}\label{3}
	\small
	\begin{aligned}
		{\theta}(s)=-\mathcal{L}^{-1}[\mathcal{N}[{\theta}({s})]]={\theta}(0)+\frac{\mathrm{d}{\theta}}{\mathrm{d}{s}}(L){s}-\int_0^{{s}}\int_L^{\tau}\mathcal{N}(s)[{\theta}]\mathrm{d}{\xi}\mathrm{d}\tau\\
	\end{aligned}	
\end{equation}
where  $\mathcal{N}[\theta(s)]$ can be derived via Adomian polynomials $\mathcal{A}_n$:	
\begin{equation}\label{series}
	\begin{aligned}
		&\mathcal{N}[\theta(s)]=\sum_{n=0}^{\infty}A_n
	\end{aligned}	
\end{equation}
where $\mathcal{A}_n$ can be obtained in an interative manner.
All parameters remain consistent with those listed in Table~\ref{tab:parameters}, except for the addition of a uniform distributed load defined as $q_y = 0.6164~\mathrm{N/m}$ and $q_x = 0~\mathrm{N/m}$. The numerical verification of the forward model is presented in Fig.~\ref{DLr}.
\subsubsection{Axial Extensibility}\label{AE}
Axial extensibility enhances input independence control in continuum robots, increasing controllable degrees of freedom. The following closed-form expressions couple bending with axial extension. Unlike earlier cases with a single input $\Delta_F$ or $\Delta l$, this configuration requires two independent inputs, $F_1$ and $F_2$ (or equivalently $\Delta l_1$ and $\Delta l_2$).
\begin{equation}\label{11}
			\small
	\begin{aligned}
	u(s) = -\frac{\Sigma_F}{EA}=\-\frac{\Sigma_l}{2L},\ 	\theta(s) = \frac{W\Delta_F}{2EI}s=\frac{\Delta_l}{W L}s
\end{aligned}
\end{equation}
where
\begin{equation}
	\small
	\begin{aligned}
&\Sigma_l= \Delta l_1 \!+ \!\Delta l_2,\ \!\!\Delta_l= \Delta l_1\! -\! \Delta l_2,\ \!\!\Sigma_F= F_1 \!+ \! F_2,\ \!\!\Delta_F= F_1\! - \!F_2\\
	\end{aligned}
\end{equation}
\begin{figure}[]
	\vspace*{0.3cm} 
	\centering
	\includegraphics[trim=0 0 0 0, clip, width=1.01\linewidth]{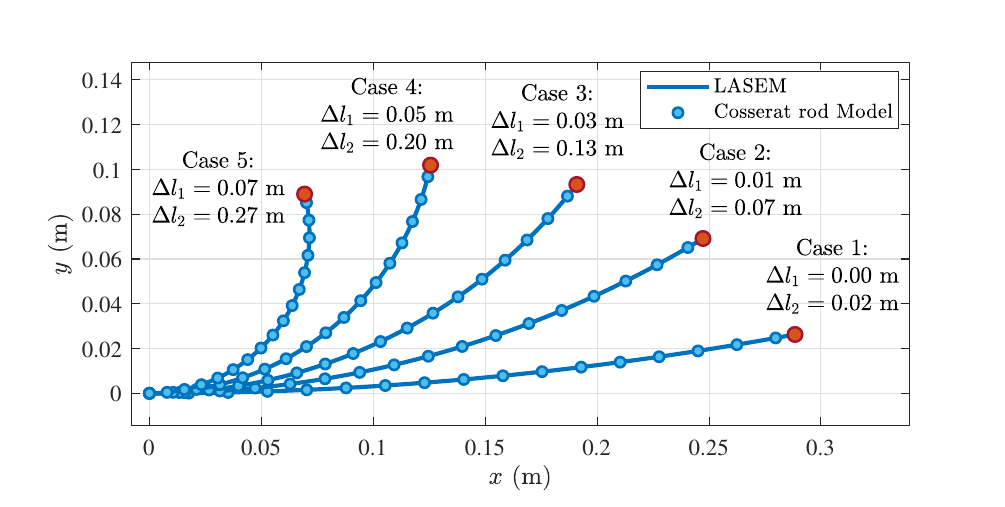}
	\caption{Forward Model Verification - Axial Extensibility}
	\label{FMV3}
\end{figure}Likewise, the relationship between the cable forces $[F_1, F_2]^\mathrm{T}$ and the cable displacement $[\Delta l_1, \Delta l_2]^\mathrm{T}$ can be derived from~\eqref{11} as:	
\begin{equation}\label{333}
	\small
	\begin{bmatrix}
		F_1 \\
		F_2
	\end{bmatrix}
	=
	\frac{1}{2}
	\begin{bmatrix}
		\frac{1}{B_2} + \frac{1}{B_1} & \frac{1}{B_2} - \frac{1}{B_1} \\
		\frac{1}{B_2} - \frac{1}{B_1} & \frac{1}{B_2} + \frac{1}{B_1}
	\end{bmatrix}
	\begin{bmatrix}
		\Delta l_1 \\
		\Delta l_2
	\end{bmatrix}
\end{equation}
where $B_1 = \frac{W^2 L}{4EI},\ B_2 = \frac{L}{EA}$. Then, following the framework of a reduced-mode Cosserat rod theory \cite{wu2023design}, we can calculate the coordinates of robot body via:
\begin{equation}\label{xu}
		\small	
	\begin{aligned}
		&x(s)=\int_0^s (1+u(s))\cos\theta(s) \mathrm{d}s,\ 
		y(s)=\int_0^s (1+u(s))\sin\theta(s) \mathrm{d}s
	\end{aligned}
\end{equation}
Finally, the closed-form inverse models are presented below via the same logic stated in Section \ref{FIK}:
\begin{equation}\label{inverse}
	\footnotesize
	\begin{aligned}
		\dot{\Delta}l_1 &= \frac{(A_3^{\Delta l} - A_4^{\Delta l})\dot{x} - (A_1^{\Delta l} - A_2^{\Delta l})\dot{y}}{2(A_2^{\Delta l} A_3^{\Delta l} - A_1^{\Delta l} A_4^{\Delta l})} \\
		\dot{\Delta}l_2 &= \frac{-(A_3^{\Delta l} + A_4^{\Delta l})\dot{x} + (A_1^{\Delta l} + A_2^{\Delta l})\dot{y}}{2(A_2^{\Delta l} A_3^{\Delta l} - A_1^{\Delta l} A_4^{\Delta l})}\\
		A_1^{\Delta l} &\!=\! -\frac{W}{2\Delta_l} \sin(\frac{\Delta_l}{W}),
		A_2^{\Delta l} \!= \!(1 - \frac{\Sigma_l}{2L}) \frac{LW}{\Delta_l}
	[ \frac{\sin(\frac{\Delta_l}{W})}{\Delta_l} + \frac{\cos(\frac{\Delta_l}{W})}{W}] \\
				A_3^{\Delta l} &\!= \!\frac{W}{2\Delta_l} [1 - \cos(\frac{\Delta_l}{W})],
		A_4^{\Delta l} = (1 \! - \! \frac{\Sigma_l}{2L}) \frac{LW}{\Delta_l}
		[ \frac{1 - \cos(\frac{\Delta_l}{W})}{\Delta_l}\!+\! \frac{\sin(\frac{\Delta_l}{W})}{W}] 
	\end{aligned}
\end{equation}
and
\begin{equation}
	\footnotesize
	\begin{aligned}
	\dot{F}_1 &= \frac{1}{4} \cdot \frac{A_1^F B_2 \dot{y} - A_2^F B_1 \dot{y} - A_3^F B_2 \dot{x} + A_4^F B_1 \dot{x}}{B_1 B_2 (A_1^F A_4^F - A_2^F A_3^F)} \\
		\dot{F}_2 &= \frac{1}{4} \cdot \frac{-A_1^F B_2 \dot{y} - A_2^F B_1 \dot{y} + A_3^F B_2 \dot{x} + A_4^F B_1 \dot{x}}{B_1 B_2 (A_1^F A_4^F - A_2^F A_3^F)}\\
		A_1^F &= -\frac{W}{4 B_2 \Delta_F} \sin( \frac{2 B_2 \Delta_F}{W} ),\\
	   A_2^F &=( 1 - \frac{B_1 \Sigma_F}{L}) \frac{L W}{2 B_2 \Delta_F}
		[ \frac{\sin( \frac{2 B_2 \Delta_F}{W})}{2 B_2 \Delta_F}
		+ \frac{\cos( \frac{2 B_2 \Delta_F}{W})}{W}] \\
		A_3^F &= \frac{W}{4 B_2 \Delta_F}[ 1 - \cos( \frac{2 B_2 \Delta_F}{W})] \\
		A_4^F &= ( 1 - \frac{B_1 \Sigma_F}{L} ) \frac{L W}{2 B_2 \Delta_F}
		[ \frac{1 - \cos( \frac{2 B_2 \Delta_F}{W} )}{2 B_2 \Delta_F}
		+ \frac{\sin( \frac{2 B_2 \Delta_F}{W} )}{W}]\\
	\end{aligned}
\end{equation}
\begin{figure}[]
	\vspace*{0.3cm} 
	\centering	
	\begin{subfigure}[b]{1.0\linewidth}
		\centering
		\includegraphics[trim=0 0 0 20, clip, width=0.94\linewidth]{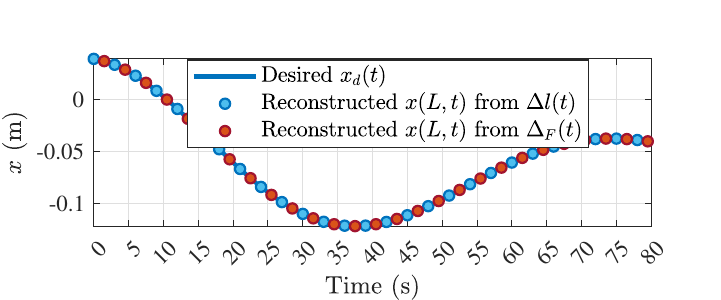}
		\caption{Inverse Model Verification: $x_d(t)$}
		\label{IKV3}
	\end{subfigure}
	\begin{subfigure}[b]{1.0\linewidth}
		\centering
		\includegraphics[trim=0 0 0 20, clip, width=0.94\linewidth]{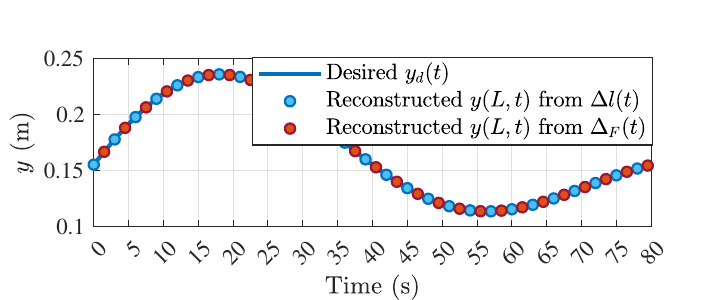}
		\caption{Inverse Model Verification: $y_d(t)$}
		\label{IKV33}
	\end{subfigure}
	\begin{subfigure}[b]{1\linewidth}
		\centering
		\includegraphics[trim=10 50 10 60, clip, width=0.98\linewidth]{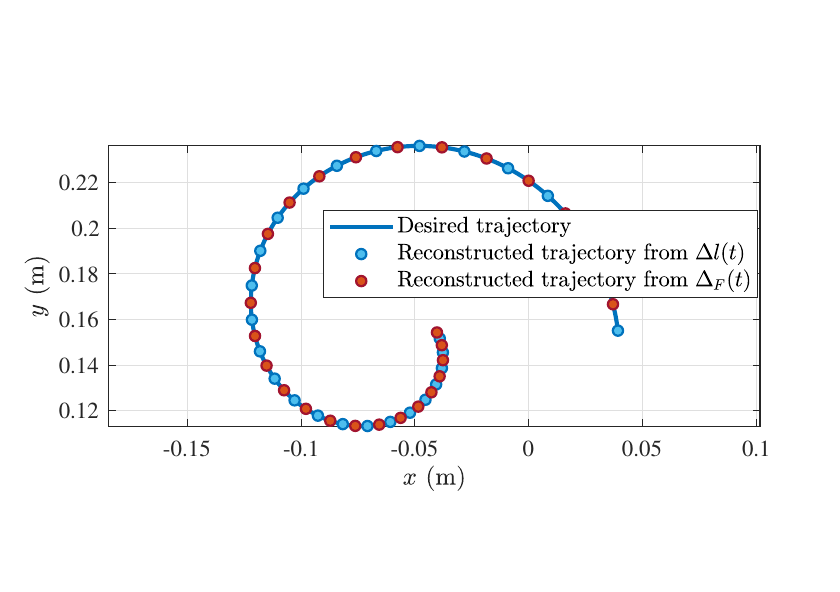}
		\caption{Inverse Model Verification: $x_d(t)$ - $y_d(t)$}
		\label{IKV34}
	\end{subfigure}
	\caption{Inverse Model Verification - Axial Extensibility}
	\label{CBr}
\end{figure}All parameters follow Table~\ref{tab:parameters}. For inverse-model verification, a circular reference trajectory with radius decreasing linearly from $R_0=0.1$\,m to $R_e=0.02$\,m is prescribed, initialized at $(x_0,y_0)$ corresponding to $\Delta l_1(0)=0.4$\,m and $\Delta l_2(0)=0.8$\,m. The time-varying radius is defined as:
\begin{equation}
	\small
	R(t) = R_0 - (R_0 - R_{e})\frac{t}{T}
\end{equation}
and the reference trajectory is given by
\begin{equation}
	\small
	\begin{aligned}
		x_d(t) = x_0 - R_0 + R(t)\cos(\frac{2\pi t}{T}),
		y_d(t) = y_0 + R(t)\sin(\frac{2\pi t}{T}
		)
	\end{aligned}
\end{equation}
The numerical verification for both forward and inverse models is illustrated in Fig.~\ref{CBr}.
\subsubsection{Summary and Analysis}
Figs.~\ref{FMV1}, \ref{FMV2}, \ref{DLr}, and \ref{FMV3} validate the proposed feedforward model against the Cosserat rod model \cite{mathew2022sorosim}, showing close agreement across all cases. Equivalent force inputs, derived from Eqs.~\eqref{111}--\eqref{333}, yield results consistent with the Cosserat model. This accuracy underscores the physical soundness of the LASEM framework while providing analytical solutions with real-time performance. To the best of our knowledge, it is the first model in the literature to achieve such completeness. For inverse modeling, Figs.~\ref{IKV}, \ref{IKV1}, \ref{IKV2}, and \ref{CBr} demonstrate accurate real-time tracking of complex trajectories using the semi-analytical inverse model, confirming the effectiveness of the LASEM framework.

\section{Considerations in Practical Implementation under LASEM framework}
\begin{figure}[]
	\vspace*{0.3cm} 
	\centering
	\includegraphics[trim=20 30 30 30, clip, scale=1.1]{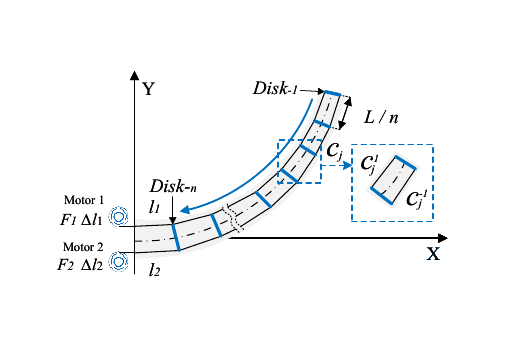}
	\caption{Schematic Diagram of a Discrete Continuum Robot}
	\label{fig:structure_diagram}
\end{figure}

The preceding derivations assume idealized models with continuously differentiable constraints. In practice, continuum robots are discretized into sequential disks \cite{walker2013continuous} (Fig.~\ref{fig:structure_diagram}), introducing effects such as discrete cable routing (Section~\ref{VCP}). Incorporating these constraints into the total potential energy $\Pi$ generally precludes analytical solutions and yields a nonlinear boundary-value problem. Classical methods, e.g., weighted residuals or Adomian decomposition (Section~\ref{DL}), provide only numerical or approximate solutions. Under the LASEM framework, we instead reformulate the functional optimization directly as a numerical optimization problem, thereby bypassing the variational step and inherently accounting for cable potential energy without explicit cable–body contact modeling.
%
\subsection{Modeling of Discrete Continuum Robots}
With the discrete disk structure in Fig.~\ref{fig:structure_diagram}, according to proposed LASEM framework, the external potential energy by the cables is accumulated span by span. Unlike Eq.~\eqref{Pi}, the total potential energy reads
\begin{equation}\label{disPi}
	\small
	\begin{aligned}
	\min_{\theta(s)} &\ \Pi = \int_0^L \frac{1}{2} EI \left( \frac{\mathrm{d}\theta}{\mathrm{d}s} \right)^2 \mathrm{d}s 
		- F_{1}\Delta l_{1}(\theta) - F_{2}\Delta l_{2}(\theta),\\
		\textbf{s.t.}\ 
		&\Delta l_{1}(\theta)
		= L - \sum_{j=1}^{n}\mathcal{C}_{j}^{(+1)}(\theta),\ 
		\Delta l_{2}(\theta)
		= L - \sum_{j=1}^{n}\mathcal{C}_{j}^{(-1)}(\theta),\\
		&\mathcal{C}_{j}^{(\sigma)}(\theta)
		= \Big[(\mathcal{A}_{j}^{(\sigma)}(\theta))^{2}
		+(\mathcal{B}_{j}^{(\sigma)}(\theta))^{2}\Big]^{1/2},\\
		&\mathcal{A}_{j}^{(\sigma)}(\theta)
		= \int_{s_{j-1}}^{s_{j}}\!\!\cos\theta(s)\mathrm{d}s
		-\sigma\,\tfrac{W}{2}\big(\sin\theta_{j}-\sin\theta_{j-1}\big),\\
		&\mathcal{B}_{j}^{(\sigma)}(\theta)
		= \int_{s_{j-1}}^{s_{j}}\!\!\sin\theta(s)\mathrm{d}s
		+\sigma\,\tfrac{W}{2}\big(\cos\theta_{j}-\cos\theta_{j-1}\big),\\
		& j=1,2,...n,\ \sigma\in\{+1,-1\},\ 
		s_0=0,\ s_j=\tfrac{jL}{n},\ \theta_j=\theta(s_j).
	\end{aligned}
\end{equation}
Here, \(n\) is the number of discrete sections (each of length \(L/n\));  \(W\) is the lateral spacing between the two symmetric cables at each disk; \(\sigma=+1\) and \(-1\) index the upper and lower cables, respectively; \(\mathcal{C}_{j}^{(\sigma)}\) is the span-wise chord length of cable \(\sigma\) across \([s_{j-1},s_j]\) computed from the integrated backbone pose with the disk-offset correction. 

The functional problem in \eqref{disPi} is generally not solvable in closed form after functional variation due to a set of nonlinear discrete constraints. To address this, as mentioned above,  we approximate the unknown function $\theta(s)$ by a polynomial expansion, justified by the Weierstrass Approximation Theorem:
\begin{equation}
	\small
	\begin{aligned}
		\theta(s) \approx \sum_{i=1}^n c_i s^i
	\end{aligned}
\end{equation}
which reduces the formulation to:
\begin{equation}\label{NumPi}
	\small
	\begin{aligned}
		\min_{c_i} \; 	\Pi=f(c_i)
	\end{aligned}
\end{equation}
This standard equality-constrained optimization problem can be solved using Lagrange multipliers. 
\subsection{{Experimental Validation}}
\begin{figure}[]
	\vspace*{0.3cm} 
	\centering
	\includegraphics[trim=0 0 0 0, clip, scale=1.1]{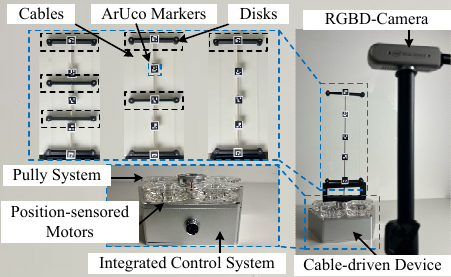}
	\caption{Experimental Setup}
	\label{fig:experimental_setup}
\end{figure}

\begin{figure}[]
	\centering
	\includegraphics[trim=0 0 0 0, clip, scale=0.65]{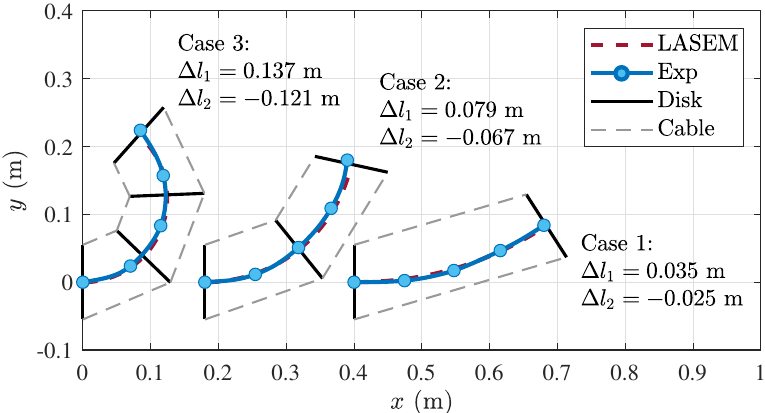}
	\caption{Experimental validation}
	\label{fig:experimental_validation}
\end{figure}

{As shown in Fig.~\ref{fig:experimental_setup}, an RGB-D camera with ArUco markers was employed to track the robot pose \cite{garrido2014automatic}, while cable displacements were applied via motor actuation. Three discretized continuum robots were evaluated: Robot~1 (left, $N=1$), 2-section Robot~2 (center, $N=2$), and 3-section Robot~3 (right, $N=3$), with geometric parameters listed in Table~\ref{tab:parameters}. Under LASEM framework, only the cable force $F_1$ was applied, with $F_1=3$, $8$, and $15$~N for $N=1$, $2$, and $3$, respectively. The corresponding cable displacements $\Delta l_1$ and $\Delta l_2$ calculated via Eq.~\eqref{disPi} were then imposed on the physical robots for validation. The experimental results exhibit strong consistency with the LASEM results for these discrete continuum robots, as demonstrated in Fig. \ref{fig:experimental_validation}. Moreover, due to discretization, the geometric constraint in Eq.~\eqref{deltal} cannot be strictly satisfied, indicating that such modeling strategies of the ideal case may not be perfectly applicable to discretized continuum robots.}
\begin{figure}[]
	\vspace*{0.3cm} 
	\centering
	\includegraphics[trim=0 0 0 0, clip, scale=0.43]{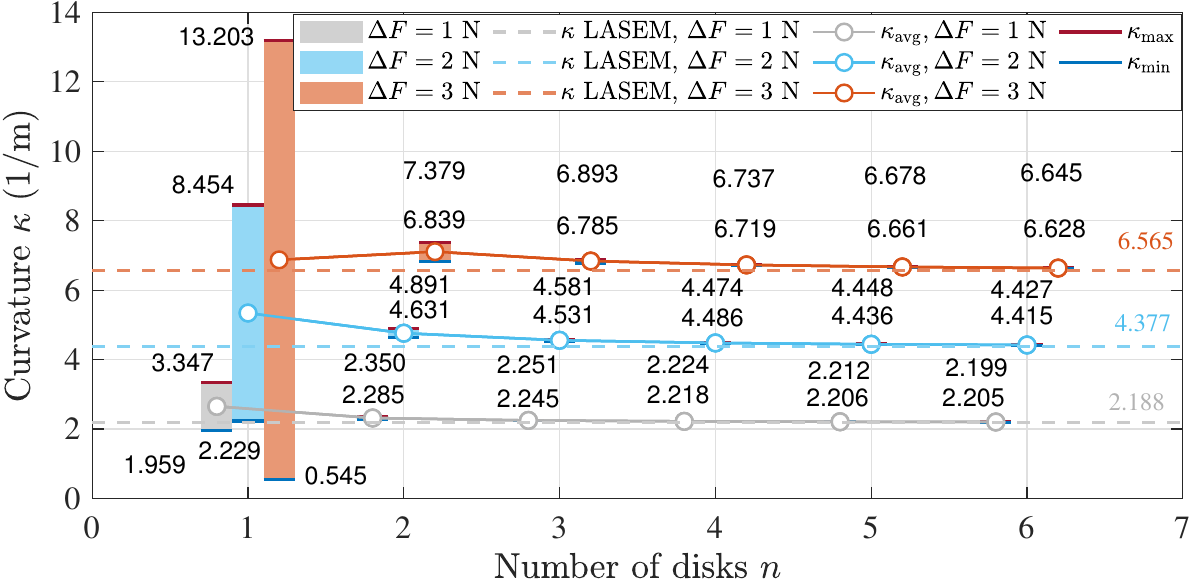}
	\caption{Discrete-to-Continuous Convergence Analysis}
	\label{fig:numerical_conv}
\end{figure}
\subsection{Convergence of Discrete to Continuous}
To investigate discretization effects on robot kinematics and statics, we analyzed the influence of the segment number $n$ in the discretized continuum robot. As shown in Fig.~\ref{fig:numerical_conv}, for a single-disk robot ($n=1$), the curvature range is wide: $\Delta F=3$~N yields $\kappa_{\min}=0.545$ and $\kappa_{\max}=13.203$, $\Delta F=2$~N yields $\kappa_{\min}=2.229$ and $\kappa_{\max}=8.454$, and $\Delta F=1$~N yields $\kappa_{\min}=1.959$ and $\kappa_{\max}=3.347$. As $n$ increases, these ranges narrow and the average curvature $\kappa_{avg}$ converges to the theoretical value from Eq.~\eqref{static} (e.g., $\kappa\approx 2.188,\,4.377,\,6.565$ for $\Delta F=1,\,2,\,3$~N, respectively). This convergence arises because larger $n$ produces cable-routing geometries that better approximate the continuous case with two inextensible cables symmetrically embedded at spacing $W=0.11$ m. These results confirm that, for sufficiently large, the discretized case converges to the analytical solution of the ideal continuous formulation under the LASEM framework.
\section{Conclusions}
This work introduced the Lightweight Actuation-Space Energy Modeling (LASEM) framework for cable-driven continuum robots. LASEM formulates actuation potential energy directly in actuation space, yielding an analytical forward model from geometrically nonlinear beam and rod theories via Hamilton’s principle, while avoiding explicit cable–backbone contact modeling. It accommodates both force and displacement inputs, unifying kinematic and static formulations, and generalizes beyond constant curvature to nonuniform geometries, arbitrary cable routings, distributed loading and axial extensibility. A semi-analytical iterative scheme was developed for inverse kinematics, and discretization was addressed by casting functional minimization as numerical optimization. Future work will extend the framework to static and dynamic modeling of spatial continuum robots also considering friction effects.

\bibliographystyle{IEEEtran}
\bibliography{reference}

\begin{thebibliography}{10}

\bibitem{walker2013continuous}
Ian~D Walker.
\newblock Continuous backbone “continuum” robot manipulators.
\newblock {\em International Scholarly Research Notices}, 2013(1):726506, 2013.

\bibitem{russo2023continuum}
Matteo Russo, Seyed Mohammad~Hadi Sadati, Xin Dong, Abdelkhalick Mohammad,
  Ian~D Walker, Christos Bergeles, Kai Xu, and Dragos~A Axinte.
\newblock Continuum robots: An overview.
\newblock {\em Advanced Intelligent Systems}, 5(5):2200367, 2023.

\bibitem{burgner2015continuum}
Jessica Burgner-Kahrs, D~Caleb Rucker, and Howie Choset.
\newblock Continuum robots for medical applications: A survey.
\newblock {\em IEEE transactions on robotics}, 31(6):1261--1280, 2015.

\bibitem{abah2021multi}
Colette Abah, Andrew~L Orekhov, Garrison~LH Johnston, and Nabil Simaan.
\newblock A multi-modal sensor array for human--robot interaction and confined
  spaces exploration using continuum robots.
\newblock {\em IEEE sensors journal}, 22(4):3585--3594, 2021.

\bibitem{wooten2018exploration}
Michael Wooten, Chase Frazelle, Ian~D Walker, Apoorva Kapadia, and Jason~H Lee.
\newblock Exploration and inspection with vine-inspired continuum robots.
\newblock In {\em 2018 IEEE International conference on robotics and automation
  (ICRA)}, pages 5526--5533. IEEE, 2018.

\bibitem{webster2010design}
Robert~J Webster~III and Bryan~A Jones.
\newblock Design and kinematic modeling of constant curvature continuum robots:
  A review.
\newblock {\em The International Journal of Robotics Research},
  29(13):1661--1683, 2010.

\bibitem{george2018control}
Thomas George~Thuruthel, Yasmin Ansari, Egidio Falotico, and Cecilia Laschi.
\newblock Control strategies for soft robotic manipulators: A survey.
\newblock {\em Soft robotics}, 5(2):149--163, 2018.

\bibitem{chirikjian1992theory}
Gregory~Scott Chirikjian.
\newblock {\em Theory and applications of hyper-redundant robotic
  manipulators}.
\newblock California Institute of Technology, 1992.

\bibitem{wang2025spirobs}
Zhanchi Wang, Nikolaos~M Freris, and Xi~Wei.
\newblock Spirobs: Logarithmic spiral-shaped robots for versatile grasping
  across scales.
\newblock {\em Device}, 3(4), 2025.

\bibitem{till2017elastic}
John Till and D~Caleb Rucker.
\newblock Elastic stability of cosserat rods and parallel continuum robots.
\newblock {\em IEEE Transactions on Robotics}, 33(3):718--733, 2017.

\bibitem{tummers2023cosserat}
Matthias Tummers, Vincent Lebastard, Fr{\'e}d{\'e}ric Boyer, Jocelyne Troccaz,
  Benoit Rosa, and M~Taha Chikhaoui.
\newblock Cosserat rod modeling of continuum robots from newtonian and
  lagrangian perspectives.
\newblock {\em IEEE Transactions on Robotics}, 39(3):2360--2378, 2023.

\bibitem{coevoet2017software}
Eulalie Coevoet, Thor Morales-Bieze, Frederick Largilliere, Zhongkai Zhang,
  Maxime Thieffry, Mario Sanz-Lopez, Bruno Carrez, Damien Marchal, Olivier
  Goury, Jeremie Dequidt, et~al.
\newblock Software toolkit for modeling, simulation, and control of soft
  robots.
\newblock {\em Advanced Robotics}, 31(22):1208--1224, 2017.

\bibitem{jones2006kinematics}
Bryan~A Jones and Ian~D Walker.
\newblock Kinematics for multisection continuum robots.
\newblock {\em IEEE Transactions on Robotics}, 22(1):43--55, 2006.

\bibitem{goury2018fast}
Olivier Goury and Christian Duriez.
\newblock Fast, generic, and reliable control and simulation of soft robots
  using model order reduction.
\newblock {\em IEEE Transactions on Robotics}, 34(6):1565--1576, 2018.

\bibitem{mathew2025reduced}
Anup~Teejo Mathew, Daniel Feliu-Talegon, Abdulaziz~Y Alkayas, Frederic Boyer,
  and Federico Renda.
\newblock Reduced order modeling of hybrid soft-rigid robots using global,
  local, and state-dependent strain parameterization.
\newblock {\em The International Journal of Robotics Research}, 44(1):129--154,
  2025.

\bibitem{armanini2023soft}
Costanza Armanini, Fr{\'e}d{\'e}ric Boyer, Anup~Teejo Mathew, Christian Duriez,
  and Federico Renda.
\newblock Soft robots modeling: A structured overview.
\newblock {\em IEEE Transactions on Robotics}, 39(3):1728--1748, 2023.

\bibitem{della2023model}
Cosimo Della~Santina, Christian Duriez, and Daniela Rus.
\newblock Model-based control of soft robots: A survey of the state of the art
  and open challenges.
\newblock {\em IEEE Control Systems Magazine}, 43(3):30--65, 2023.

\bibitem{neppalli2008geometrical}
Srinivas Neppalli, Matthew~A Csencsits, Bryan~A Jones, and Ian Walker.
\newblock A geometrical approach to inverse kinematics for continuum
  manipulators.
\newblock In {\em 2008 IEEE/RSJ International Conference on Intelligent Robots
  and Systems}, pages 3565--3570. IEEE, 2008.

\bibitem{dickson2025real}
Akua Dickson, Juan C~Pacheco Garcia, Ran Jing, Meredith~L Anderson, and
  Andrew~P Sabelhaus.
\newblock Real-time trajectory generation for soft robot manipulators using
  differential flatness.
\newblock In {\em 2025 IEEE 8th International Conference on Soft Robotics
  (RoboSoft)}, pages 1--7. IEEE, 2025.

\bibitem{renda2014dynamic}
Federico Renda, Michele Giorelli, Marcello Calisti, Matteo Cianchetti, and
  Cecilia Laschi.
\newblock Dynamic model of a multibending soft robot arm driven by cables.
\newblock {\em IEEE Transactions on Robotics}, 30(5):1109--1122, 2014.

\bibitem{mathew2022sorosim}
Anup~Teejo Mathew, Ikhlas~Ben Hmida, Costanza Armanini, Frederic Boyer, and
  Federico Renda.
\newblock Sorosim: A matlab toolbox for hybrid rigid--soft robots based on the
  geometric variable-strain approach.
\newblock {\em IEEE Robotics \& Automation Magazine}, 30(3):106--122, 2022.

\bibitem{buss2005selectively}
Samuel~R Buss and Jin-Su Kim.
\newblock Selectively damped least squares for inverse kinematics.
\newblock {\em Journal of Graphics tools}, 10(3):37--49, 2005.

\bibitem{wu2023design}
Ke~Wu.
\newblock {\em Design, Modeling, Optimization and Control of Compliant
  Mechanisms}.
\newblock PhD thesis, Centrale Lille Institut, 2023.

\bibitem{renda2022geometrically}
Federico Renda, Costanza Armanini, Anup Mathew, and Frederic Boyer.
\newblock Geometrically-exact inverse kinematic control of soft manipulators
  with general threadlike actuators’ routing.
\newblock {\em IEEE Robotics and Automation Letters}, 7(3):7311--7318, 2022.

\bibitem{garrido2014automatic}
Sergio Garrido-Jurado, Rafael Mu{\~n}oz-Salinas, Francisco~Jos{\'e}
  Madrid-Cuevas, and Manuel~Jes{\'u}s Mar{\'\i}n-Jim{\'e}nez.
\newblock Automatic generation and detection of highly reliable fiducial
  markers under occlusion.
\newblock {\em Pattern Recognition}, 47(6):2280--2292, 2014.

\end{thebibliography}

\end{document}